\documentclass{article}

\usepackage{microtype}
\usepackage{graphicx}
\usepackage{subfigure}
\usepackage{booktabs} 

\usepackage{hyperref}

\usepackage[accepted]{mlsys2020}

\usepackage{amsmath,amssymb}

\usepackage[utf8]{inputenc} 
\usepackage[T1]{fontenc}    
\usepackage{amsfonts}       
\usepackage{nicefrac}       

\usepackage{tikz}
\usepackage{blindtext}
\usepackage{url}
\usepackage{float}
\usepackage{multirow}
\usepackage{caption}
\usepackage{comment}
\usepackage[titlenumbered,ruled,algo2e]{algorithm2e}
\renewcommand{\algorithmiccomment}[1]{\hfill{$\triangleright$ #1}}
\usepackage{array}
\usepackage{wrapfig}
\usepackage{fdsymbol}
\usepackage{authblk}

\usepackage{bbm}

\newcommand{\PaperTitle}{DNA: Differentiable Network-Accelerator Co-Search}

\definecolor{Note_color}{rgb}{1.0, 0.0, 0.0}

\mlsystitlerunning{\PaperTitle}

\begin{document}

\twocolumn[
\mlsystitle{\PaperTitle}



\mlsyssetsymbol{equal}{*}

\begin{mlsysauthorlist}
\mlsysauthor{Yongan Zhang}{equal,rice}
\mlsysauthor{Yonggan Fu}{equal,rice}
\mlsysauthor{Weiwen Jiang}{nd}
\mlsysauthor{Chaojian Li}{rice}
\mlsysauthor{Haoran You}{rice}
\\
\mlsysauthor{Meng Li}{fb}
\mlsysauthor{Vikas Chandra}{fb}
\mlsysauthor{Yingyan Lin}{rice}
\end{mlsysauthorlist}

\mlsysaffiliation{rice}{Department of Electrical and Computer Engineering, Rice University}
\mlsysaffiliation{fb}{Facebook, Inc}
\mlsysaffiliation{nd}{Department of Computer Science and Engineering, University of Notre Dame}

\mlsyscorrespondingauthor{Yingyan Lin}{yingyan.lin@rice.edu}

\mlsyskeywords{Machine Learning, MLSys}

\vskip 0.3in

\begin{abstract}
Powerful yet complex deep neural networks (DNNs) have fueled a booming demand for efficient DNN solutions to bring DNN-powered intelligence into numerous applications. 
Jointly optimizing the networks and their accelerators are promising in providing optimal performance. However, the great potential of such solutions have yet to be unleashed due to the challenge of \textit{simultaneously} exploring the vast and entangled, yet different design spaces of the networks and their accelerators.
To this end, we propose DNA, a \textbf{D}ifferentiable \textbf{N}etwork-\textbf{A}ccelerator co-search framework for automatically searching for matched networks and accelerators to maximize both the task accuracy and acceleration efficiency.  
Specifically, DNA integrates two enablers: (1) \textit{a generic design space for DNN accelerators} that is applicable to both FPGA- and ASIC-based DNN accelerators and compatible with DNN frameworks such as PyTorch to enable algorithmic exploration for more efficient DNNs and their accelerators; 
and (2) \textit{a joint DNN network and accelerator co-search algorithm} that enables the simultaneous search for optimal DNN structures and their accelerators' micro-architectures and mapping methods to maximize both the task accuracy and acceleration efficiency. 
Experiments and ablation studies based on FPGA measurements and ASIC synthesis show that the matched networks and accelerators generated by DNA consistently outperform state-of-the-art (SOTA) DNNs and DNN accelerators (e.g., $3.04\times$ better FPS with a $5.46\%$ higher accuracy on ImageNet), while requiring notably reduced search time (up to $1234.3\times$) over SOTA co-exploration methods, when evaluated over ten SOTA baselines on three datasets.
\end{abstract}

]



\printAffiliationsAndNotice{\mlsysEqualContribution} 

\begin{figure*}[!ht]
  \vspace{-0.2cm}
    \centering
    \includegraphics[width=\linewidth]{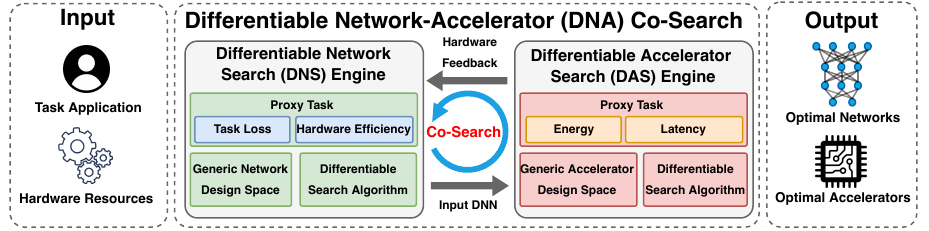}
         \vspace{-1.5em}
    \caption{An illustration of our DNA co-search framework, which accepts target tasks and accelerator specifications and then automatically generates matched DNNs and their accelerators to maximize both task accuracy and hardware efficiency.}
    \label{fig:overview}
     \vspace{-1em}
\end{figure*}

\vspace{-0.6cm}
\section{Introduction}
\label{intro}
 
Powerful deep neural networks (DNNs)' prohibitive complexity stands at odds with the limited resources of daily life devices and has raised various environmental concerns \cite{Strubell2019Energy}, motivating intensive studies of efficient DNN solutions. Early works merely explore from either the algorithm or hardware level. For example, compression techniques attempt to trim down DNNs' complexity, while on the hardware level representative FPGA- and ASIC-based accelerators \cite{chen2016eyeriss,du2015shidiannao,zhang2018dnnbuilder} develop customized \textit{micro-architectures} (e.g., \# of memory hierarchies and processing elements (PEs), PE array dimension and shape, size of different memories, and network-on-chip (NoC) design) and algorithm-to-hardware \textit{mapping methods} (e.g., loop tiling strategy, loop size, and loop order) to boost DNN acceleration efficiency. 
Later, hardware-aware neural architecture search (HA-NAS) \cite{cai2018proxylessnas,wu2019fbnet,tan2019mnasnet} emerged to automate the design of efficient \textit{network structures} (e.g., \# of layers and channels, size of kernels, and layer operations). Recently, it has been recognized that maximizing DNN accelerators’ efficiency requires joint exploration of both the networks and their accelerators (the latter refers to the accelerators' micro-architectures and mapping methods hereafter) ~\cite{edge_fpga_co_design, abdelfattah2020best,li2020edd,Lin2019NeuralHardwareAS}.

Despite their promise, existing works
have yet to unleash the great potential of jointly optimizing DNNs and their accelerators. 
The key challenges include (1) the \textit{prohibitively large} joint space consisting of the \textit{coupled yet different} network and accelerator spaces with extremely sparse optima, (2) \textit{non-differentiable} hardware costs, and (3) how to \textit{algorithmically describe} a generic accelerator search space. 
We aim to tackle the aforementioned challenges, and make the following contributions:

\begin{itemize}
  \vspace{-1em}

\item We propose \textbf{DNA}, a \textbf{D}ifferentiable \textbf{N}etwork-\textbf{A}ccelerator co-search framework (see Fig.~\ref{fig:overview}) that enables a \textbf{joint search of DNNs' network structures and their accelerators' micro-architectures and mapping methods}, promising to largely boost the efficiency and expedite the development of DNN accelerators.

\vspace{-0.5em}

\item We develop Differentiable Accelerator Search (DAS), a generic differentiable accelerator search engine for exploring the large and discrete design space of both DNN accelerators' micro-architectures and mapping methods. DAS 
distinguishes itself from existing accelerator searches which mostly adopt RL-based methods and thus are limited in scalability and performance.
 
\item We construct a Generic DNN Accelerator Design Space (GADS) that is applicable to different DNN accelerators including both FPGA- and ASIC-based ones. Such a GADS can serve as a key enabler for algorithmically exploring both (1) DNN accelerators' large and discrete design space and (2) hardware-driven efficient DNNs.

\item Through FPGA measurements and ASIC synthesis, extensive experiments and ablation studies validate DNA's effectiveness and superiority: DNA generated networks/accelerators consistently outperform SOTA DNNs/accelerators, while requiring a notably reduced search time compared to SOTA co-exploration methods. 
We also visualize DNA generated accelerators to provide insights towards efficient DNN accelerators. 

\end{itemize}

  \vspace{-0.3cm}
\begin{figure}[tbh]
    \centering
    \includegraphics[width=\linewidth]{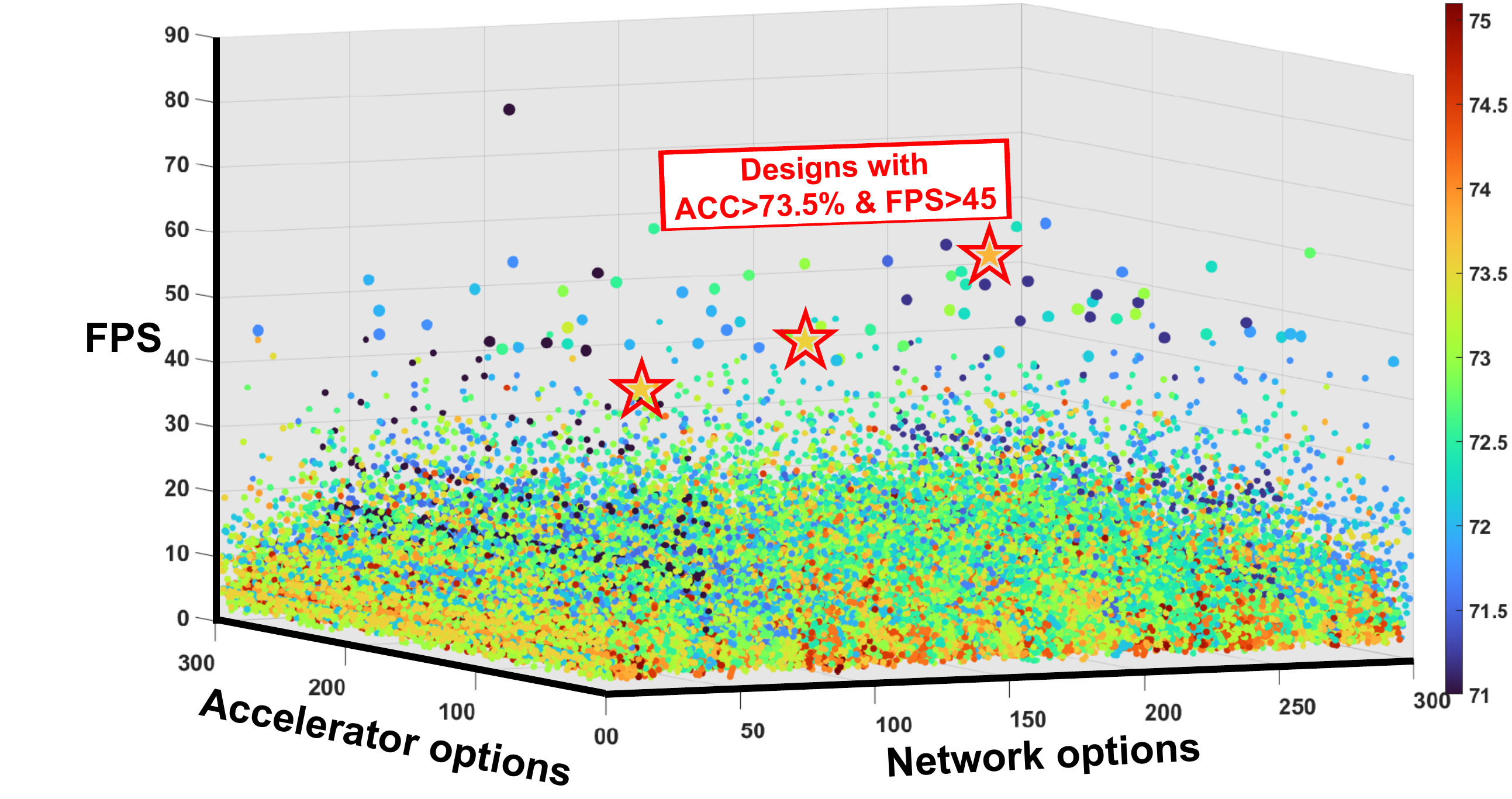}
    \caption{FPGA measured Frame-Per-Second (FPS; see the left axis) on a ZC706 FPGA~\cite{zc706} and CIFAR-100 based accuracy (see the right colorbar) of 300 randomly sampled networks from the FBNet~\cite{wu2019fbnet} search space, when each of the networks is accelerated by 300 randomly sampled accelerators from a generic accelerator design space, leading to a total of $9$E+4 randomly sampled data points in this figure. Designs with $ACC>73.5\%$ and $FPS >45$ are marked as stars, which are extremely sparse in the search space. }
    \label{fig:huge_space}
    \vspace{-1.5em}
\end{figure}
  
\section{Related Works}
\textbf{Hardware-Aware NAS (HA-NAS).} 
HA-NAS has been developed to automate the design of efficient DNNs. Early works ~\cite{tan2019mnasnet, howard2019searching, tan2019efficientnet} utilize RL-based methods, and thus suffer from substantial search time and costs, limiting their scalability. Later, motivated by DARTS~\cite{liu2018darts}, differentiable HA-NAS~\cite{wu2019fbnet, wan2020fbnetv2, jin2019rc, li2020edd} emerged to greatly improve both the search and hardware efficiency, respectively. 
However, existing HA-NAS methods (1) mostly consider hardware costs (e.g., FLOPs or latency) on one given device/accelerator and (2) have not yet fully explored the hardware design space. For maximizing DNNs' acceleration efficiency determined by both the networks and their underlying hardware, it is highly desired to jointly search for both the networks and their hardware accelerators.  


\textbf{DNN accelerators.} 
DNNs' powerful performance and prohibitive complexity have motivated extensive research in customized DNN accelerators. Given a DNN and its acceleration specification, SOTA accelerators \cite{du2015shidiannao,chen2017eyeriss,zhao2020smartexchange,li2020timely} explore different micro-architectures and algorithm-to-hardware mapping methods to maximize data reuses and thus acceleration efficiency. Early works rely on experts' manual design, which can be very time consuming and require cross-disciplinary \sloppy knowledge in algorithm, micro-architecture, and circuit design. Recently, there has been a growing interest in design flow~\cite{vivado_HLS,hls_chen2005xpilot,hls_chen2009lopass,hls_rupnow2011high} and DNN accelerator design automation ~\cite{wang2016deepburning, zhang2018caffeine, guan2017fp, venkatesanmagnet,wang2018design}. However, these works mostly explore the accelerator design space, leading to sub-optimal solutions.

\begin{figure*}[tbh]
    \centering
    \includegraphics[width=\linewidth]{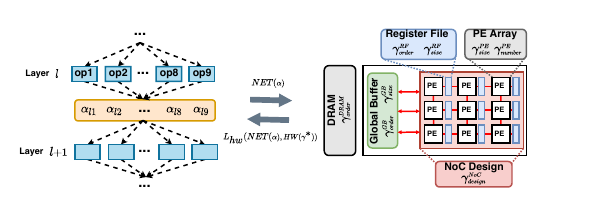}
         \vspace{-4em}
         \caption{DNA's differentiable co-search of the network and accelerator joint space, where the DAS engine (right) optimizes the accelerator parameters based on Eq.~\ref{eq:update_hw} and Eq.~\ref{eqn:obj_hw} based on the input networks $NET(\alpha)$ and returns the corresponding hardware cost to the DNS engine (left) for which to penalize the costly operators. Here $\gamma^{rf}_{order}$ denotes the accelerator parameter of \textit{loop-order} in the register file (RF), and similar notations are adopted for other accelerator parameters in Tab.~\ref{tab:hw_space}.
    }
    \label{fig:co-search}
     \vspace{-1.2em}
\end{figure*}

\textbf{Software/Hardware co-exploration.} 
Jointly exploring the networks and their accelerators has been shown to be promising~\cite{edge_fpga_co_design, abdelfattah2020best, yang2020co, jiang2020device, jiang2019hardware, li2020edd}. 
For example,~\cite{edge_fpga_co_design, abdelfattah2020best,jiang2019hardware} conduct RL-based search to jointly search for the networks and design parameters of an FPGA-based accelerator;~\cite{yang2020co} adopts RL-based controllers to search for the networks and to determine which of the two ASIC accelerators should be selected; and~\cite{li2020edd,choi2020dance} extend differentiable NAS to network and accelerator co-search.

Despite their promise, there is still much room to further enlarge the extent of co-exploration and thus improve the achieved performance. First, existing works have not yet considered a generic accelerator space including both the micro-architectures 
(e.g., \# of memory hierarchies/PEs, PEs' dimension/shape, size of different memories, and NoC design) and the mapping methods (e.g., loop tiling strategy, and loop size/order), which is also \textbf{a key challenge for co-search/exploration}. Second, the RL-based methods ~\cite{edge_fpga_co_design, abdelfattah2020best, yang2020co, jiang2020device, kang2018memory} involve large search costs, limiting their \textbf{scalability} to handle the large joint space, and thus only a very limited accelerator space (e.g., two accelerator choices~\cite{yang2020co}) is considered. Despite EDD's pioneering effort to differentiably search for both the network structures and their precision, it only considers one accelerator factor (i.e., the parallel factor of an FPGA accelerator template) which can be analytically fused into their theoretical computational cost, and this is not always applicable to most naturally non-differentiable accelerator design parameters such as loop order and loop size. A concurrent work~\cite{choi2020dance} adopts a DNN to generate suggested accelerator designs based on the network structures that serve as the DNN's inputs. While simple, such a training based method lacks interpretability; and 
they only evaluate on an Eyeriss~\cite{chen2016eyeriss} based template without benchmarking over SOTA co-exploration works, leaving their general effectiveness unclear.

\vspace{-0.2cm}
\section{The proposed DNA framework}
\label{sec:proposed}
\vspace{-0.2cm}
This section describes our DNA framework. We first provide an overview and the problem formulation, and then DNA's co-search algorithm, followed by DNA's two search engines and the proposed generic accelerator search space.

\begin{figure*}[!t]
    \begin{minipage}{\textwidth}
    \begin{algorithm}[H]
    \DontPrintSemicolon
    \KwIn{supernet weight $\omega$, the network space $NET(\alpha)$, the accelerator space $HW(\gamma)$, the total search epoch $max\_epoch$}
    
    \KwOut{the optimal network  $optimal\_net$ and the optimal accelerator $optimal\_accelerator$}

   \While{epoch $< max\_epoch$ }
   {
     Obtain optimized accelerators $HW_m^* (m=1,...,M)$ using Eq.~\ref{eqn:obj_hw} for $M$ DNNs sampled from the current network distribution $NET(\alpha)$ and calculate the average hardware cost for each operator based on Eq.~\ref{eqn:expect}
     
     \For{one training epoch}
     {
        update $\omega$ based on Eq.~\ref{eq:update_weight}
        
        update $\alpha$ based on Eq.~\ref{eq:update_alpha} where $L_{hw}$ is calculated using Eq.~\ref{eqn:optim_eff}
     }
   }
   
   Derive $optimal\_net$ by selecting the operator with the maximal weighted coefficients for each layer based on Eq.~\ref{eqn:arch_loss} 
   
   Derive $optimal\_accelerator$ for the optimal $optimal\_net$ based on Eq.~\ref{eqn:obj_hw} 
   
   \Return\{{$optimal\_net$, $optimal\_accelerator$}\}
 
    \caption{DNA's differentiable network-accelerator co-search algorithm.} \label{alg:dna}
    \end{algorithm}
\end{minipage}
\vspace{-1em}
\end{figure*}

\vspace{-0.2cm}
\subsection{DNA: overview and formulation}
\label{sec:overview}
As mentioned in Sec. 1, the challenges of effective network-accelerator co-search include (1) the \textbf{prohibitively large and irregular} joint space versus \textbf{very sparse} optima excelling at both accuracy and efficiency, as shown in Fig.~\ref{fig:huge_space}, (2) non-differentiable hardware costs, and (3) the lack of a generic accelerator search space description.


Our DNA framework aims to tackle the first of the aforementioned challenges via gradient-based 
optimization to effectively and efficiently identify the optimal network and accelerator pairs.
As shown in Fig.~\ref{fig:overview}, DNA accepts user-defined datasets, target performance (e.g., task accuracy and hardware efficiency), and resource constraints as inputs to automatically generate matched pairs of DNNs and their accelerators to maximize both the accuracy and efficiency given the resource constraints. DNA consists of two search engines: (1) the DAS engine integrated with a generic accelerator design space, which enables differentiable search to handle the accelerators' \textit{large} and \textit{discrete} design space, and (2) the DNS engine, built upon SOTA differentiable NAS works~\cite{wu2019fbnet}. Formally, we can formulate DNA's optimization as:
  \vspace{-0.1cm}
\begingroup
\allowdisplaybreaks
\begin{align} 
    \begin{split}
    & \min \limits_{\alpha} \,\, L_{val}( \omega^*, NET(\alpha))+\lambda L_{hw}(NET(\alpha), HW(\gamma^*)) \label{eq:update_alpha} 
    \end{split} \\
    \begin{split}
    & s.t. \quad \omega^* = \underset{\omega}{\arg\min} \,\, L_{train}( \omega,  NET(\alpha)),  \label{eq:update_weight}
    \end{split} \\
    \begin{split}
    & s.t. \quad \gamma^* = \underset{\gamma}{\arg\min} \,\, L_{hw}(NET(\alpha), HW(\gamma)) \label{eq:update_hw}
    \end{split}
\end{align}
\endgroup
where $L_{train}$ and $L_{val}$ are the task loss of training and validation, respectively; $\omega$, $\alpha$, and $\gamma$ are the supernet weights \cite{liu2018darts}, DNNs' structure parameters (i.e., architecture parameters in~\cite{liu2018darts}), and the accelerator parameters, respectively; $NET(\alpha)$ and $HW(\gamma)$ denote the network and the accelerator space parameterized by $\alpha$ and $\gamma$, respectively; and $L_{hw}$ is the hardware-cost loss determined by both the network and its accelerator.

Despite its simplicity in formulation, it is in general intractable to analytically solve Eq.~\ref{eq:update_alpha}$\sim$Eq.~\ref{eq:update_hw}.
DNA integrates its DNA and DAS engines to simultaneously search for optimally matched networks and accelerators at much improved search efficiency, as illustrated in Fig.~\ref{fig:co-search}.



\subsection{DNA: the joint search algorithm}
\label{sec:DNA}
Here we describe DNA's co-search algorithm. There are two practical challenges to design a differentiable co-search algorithm for effectively navigating through the network and accelerator joint space. First, ideally the hardware-cost penalty for each layer-wise network operator should be obtained when being executed on the final searched optimal accelerator, which is not yet available at each co-search epoch as the optimal network is still unknown, i.e., an chicken-and-egg problem. Second,  the network is searched by regularizing $\alpha$ in a layer-wise manner (see Eq.~\ref{eqn:arch_loss}), while the accelerator $\gamma$ is determined by the whole DNN (see Eq.~\ref{eqn:obj_hw}) due to the global accelerator parameters shared among all the layers, e.g., the loop-order in Tab.~\ref{tab:hw_space}.  


To address these challenges, DNA obtains the hardware-cost loss $L_{hw}$ at each co-search epoch by approximating the final optimal accelerator using the accelerator optimized for the network consisting of layer-wise operators that associate with a high probability.
The hypothesis is that the network operators that have higher probabilities are also more likely to appear in the final optimal network, and thus, the corresponding optimal accelerators are likely to be the final optimal one. Specifically, at each epoch, DNA samples $M$ networks from the current network distribution $NET(\alpha)$ and obtains the optimal accelerator for each of them using its DAS engine (see Sec~\ref{sec:DAS}); the hardware-cost loss of each operator is then obtained using its average hardware cost on the $M$ optimal accelerators generated from the previous step. For example, the hardware cost of the $k$-th operator in the $l$-th layer $O_{lk}$ is formulated as:


\vspace{-2em}

\begin{equation}
\begin{aligned}
    \hspace{3em} E_{NET\sim P(NET|\alpha)}&( L_{hw}(O_{lk}, HW(\gamma^*))) \\
    \approx \frac{1}{M}\sum_{m=1}^{M} L&_{hw}(O_{lk},HW_m^*) 
\end{aligned}
\label{eqn:expect}
\vspace{-0.5em}
\end{equation}

\vspace{-1em}

where $HW_m^*$ is the optimal accelerator generated using DNA's DAS engine (see Eq.~\ref{eqn:obj_hw} in Sec.~\ref{sec:DAS}) for the $m$-th sampled network, and $L_{hw}(O_{lk},HW_m^*)$ is the hardware-cost loss of accelerating the operator $O_{lk}$ using the accelerator $HW_m^*$. 
As a result, the hardware-cost loss of the whole DNN in Eq.~\ref{eq:update_alpha} can be formulated in a layer-wise manner:

\vspace{-1.5em}

\begin{equation}
\begin{aligned}
&L_{hw}(NET(\alpha), HW(\gamma^*)) \\
&= \sum_{l=1}^{L}\sum_{k=1}^{K} \alpha_{lk} {E}_{NET\sim P(NET|\alpha)}( L_{hw}(O_{lk}, HW(\gamma^*))) \\
&\approx \frac{1}{M} \sum_{l=1}^{L} \sum_{k=1}^{K} \sum_{m=1}^{M} \alpha_{lk} L_{hw}(O_{lk}, HW_m^*)
\end{aligned}
\label{eqn:optim_eff} 
\end{equation}

\vspace{-1em}

Essentially, the hardware-cost loss $L_{hw}$ at each co-search epoch is approximated using the weighted sum of the approximated hardware cost for all the layer-wise operators.
The co-search algorithm of DNA is summarized in Alg.~\ref{alg:dna}, the effectiveness and superiority of which are consistently validated in our experiments (see Sec.~\ref{sec:result}). For example, DNA boosts the search efficiency by orders-of-magnitude while leading to superior accuracy and hardware efficiency.

\subsection{DNA: the DNS engine}
\label{sec:DNS}
\vspace{-0.3em}

The DNS engine can be realized by leveraging SOTA differentiable NAS works. First, for the network search space, we consider the SOTA hardware-friendly search space in~\cite{wu2019fbnet} which searches the kernel size, channel expansion ratio, and group number for each building block; Second, for the network search algorithm, we adopt the SOTA differentiable one with Gumbel-Softmax in~\cite{wu2019fbnet}, which computes the output of the $l$-th layer $A_l$ as a weighted sum of all candidate operators: 
\begin{equation}
\label{eqn:arch_loss}
    A_l = \sum_{k=1}^{K} \alpha_{lk} O_{lk}(A_{l-1})
\end{equation}
where $K$ denotes the total number of layer-wise candidate operators, $O_{lk}$ denotes the $k$-th operator for the $l$-th layer, and $\alpha_{lk}$ denotes  the weighted coefficient of $O_{lk}$.

\vspace{-0.3em}
\subsection{DNA: the DAS engine}
\label{sec:DAS}
\vspace{-0.2em}
EDD~\cite{li2020edd} attempts a pioneering step to differentiably co-search the network and its accelerator, their search space is yet limited to include only one accelerator parameter (i.e., the parallel factor) within their accelerator template. Their parallel factor can be analytically fused into the theoretical computational cost (e.g., the number of FLoating-point OPerations or FLOPs) in their framework. 
However, such a fuse strategy is not always applicable to naturally non-differentiable accelerator parameters such as the loop size and loop order, which are critical to the hardware efficiency \cite{chen2016eyeriss,venkatesanmagnet}. A more general and efficient accelerator search engine is thus highly desired to unleash the potential of network-accelerator co-search.


Our DAS engine aims to close the above gap and realizes a generic differentiable accelerator search engine built on top of our GADS (see Sec.~\ref{sec:space}). Specifically,  
we reformulate Eq.~\ref{eq:update_hw} and propose a differentiable method to solve it: 
 \vspace{-0.1em}
\begin{equation} \label{eqn:obj_hw}
\resizebox{\linewidth}{!}{
   $ \gamma^* = \,\, \underset{\gamma}{\min} \,\, \sum_{s=1}^{S} GS(\gamma^s) \, L_{hw}(NET(\alpha^*), HW(GS(\gamma^1), ..., GS(\gamma^S)))$\\
    }
\end{equation}
where the accelerator $HW$ is characterized by its accelerator parameters ${\gamma^s \; (s=1,...,S)}$, which is a normalized vector representing the $s$-th accelerator parameter with each element of $\gamma^s$ defining the probability of the corresponding choice of its represented accelerator parameter, and $GS(\gamma^s)$ denotes Gumbel-Softmax sampling~\cite{gumbel1948statistical, maddison2014sampling} of the $s$-th accelerator parameter $\gamma^s$.

Unlike NAS, different options of one accelerator parameter are NOT additive, i.e., cannot be formulated as a sum weighted by the probability as in~\cite{liu2018darts}. As such, for each accelerator parameter, we apply Gumbel-Softmax sampling~\cite{gumbel1948statistical, maddison2014sampling} to sample only one choice $GS(\gamma^s)$ of the $s$-th accelerator parameter. Once all the accelerator parameters are sampled, the corresponding DNN accelerator's hardware efficiency can be obtained using SOTA accelerator performance estimators, where in this work we refer to ~\cite{xu2020autodnnchip} for FPGA-based accelerators and ~\cite{wu2019accelergy,parashar2019timeloop} for ASIC-based accelerators.
We then multiply the resulting hardware-cost loss with the sampled $GS(\gamma^s)$ and relax to Gumbel-Softmax~\cite{jang2016categorical} during backpropagation for estimating the gradients.

\subsection{Generic Accelerator Design Space (GADS)}
\label{sec:space}
  \vspace{-0.2cm}

Similar to NAS, a generic accelerator search space is a prerequisite for algorithmic accelerator exploration and optimization. However, it is challenging to develop such a space for DNN accelerators due to their \textit{large} and \textit{discrete} design space. First, there are numerous choices for the algorithm-to-hardware \textbf{mapping methods} (i.e., how to \textit{temporally} and \textit{spatially} schedule all the DNN's operations to be executed in the target accelerators).
Second, there are many ways to design the accelerators' \textbf{micro-architectures}, which are characterized by the number of memory hierarchies and PEs, the size of each memory hierarchy, the shape and size of the PE array, and the NoC design~\cite{chen2017eyeriss}. 


\begin{table}[!h]
  \centering
  \caption{The constructed generic accelerator search space, where TBS denotes ``to be searched''.}
      \vspace{-0.8em}
    \resizebox{0.5\textwidth}{!}{
        \begin{tabular}{ccc}
        \toprule
        \textbf{Memory Hierarchy} & \textbf{Loop-order} & \textbf{Loop-size} \\
        \midrule
        \textbf{DRAM}   & TBS    & \multicolumn{1}{c}{-} \\
        \textbf{Global Buffer} & TBS    & TBS \\
        \textbf{PE array}    & \multicolumn{1}{c}{-} & TBS \\
        \textbf{Register File (RF)} & TBS    & TBS \\
        \midrule
        \midrule
        \textbf{NoC design} & \textbf{Max \# of PEs} & \textbf{Pipeline/Multi-cycle} \\
        \midrule
        TBS    & TBS    & TBS \\
        \bottomrule
        \end{tabular}%
    }
  \label{tab:hw_space}%
  \vspace{-1em}
\end{table}

We construct a generic accelerator search space as shown in Tab.~\ref{tab:hw_space} by leveraging the commonly used nested \textit{for-loop} accelerator description~\cite{chen2016eyeriss,parashar2019timeloop,blocking_cnn,zhang2015optimizing,zhao2020icassp} which naturally bridges the accelerator's micro-architectures and mapping methods with DNNs' network parameters. More details about the nested \textit{for-loop} description can be found in the appendix. Next, we introduce each accelerator parameter in Tab.~\ref{tab:hw_space}:

\textbf{\textit{loop-order}}: the orders of the loops within each memory hierarchy, each of which has a total of $n$ data dimensions. As such, $n$ loops correspond to an $n$-item ordering problem. To be compatible with the proposed DAS engine in Sec.~\ref{sec:DAS}, where each accelerator parameter should have all possible choices parameterized by the corresponding $\gamma$ vector (see Eq.~\ref{eqn:obj_hw}), we formulate the \textit{loop-order} as a problem of picking one choice from a total of $n$ options without replacement for $n$ times (e.g., $n=6$ considering the number of data dimensions in DNNs).

\textbf{\textit{loop-size}}:
the size of each loop in the \textit{for-loop} description. The product of all loop-sizes associated with each data dimension needs to equal the corresponding algorithmic dimension, because the nested loop sizes as a whole dictate the total number of execution iterations. Then, intuitively, the possible choices for a certain loop's size are all the choices that the corresponding data dimension can be factorized into. 


\textbf{\textit{NoC design}}: the parallel execution pattern of MACs (multiply–accumulate operations) when accelerating DNNs on an accelerator, which is determined by the PE array style. In this work, we consider three NoC options following the common practice, as inspired by SOTA accelerators~\cite{chen2016eyeriss,zhang2015optimizing,zhao2020icassp}: 
\vspace{-1em} 
\begin{itemize}
\setlength{\itemsep}{0pt}
    \item parallelizing the computation over the output partial sums, where the dimensions of output channels, output rows, and output columns are executed in parallel.
    \item parallelizing the computation over the kernels, where the dimensions of output channels, input channels, kernel rows, and kernel columns are executed in parallel.
    \item parallelizing the computation over both the kernel and output dimensions, where the dimensions of output channels, kernel rows, and output columns are executed in parallel.
\end{itemize}
\vspace{-1em}

\textbf{\textit{max number of PEs}}: the maximal number of PEs in the design which can range from 1 to a specified value 
and are determined by the area constraint. 

\textbf{\textit{pipeline/multi-cycle}}: a binary choice between a chunk-based pipeline  micro-architecture of the DNN accelerator and a multi-cycle micro-architecture shared by all layers, inspired by SOTA FPGA-based accelerators~\cite{zhang2018dnnbuilder,shen2017ISCA}, with each of the two choices having different designs. For example, if the chunk-based pipeline micro-architecture is chosen, the allocation of different DNN layers to different chunks would be a hyperparameter and often differ for different DNN structures and accelerators. In our DAS, each layer's assignment can be formulated and parameterized by a $\gamma$ vector. Also, depthwise convolutions are assigned to a different set of chunks considering their unique structures as compared to vanilla convolutions. 

\vspace{-0.2cm}
\section{Experiment results}
\label{sec:result}
In this section, we first introduce the experiment setup, and then evaluate DNA over both SOTA (1) HW/SW co-exploration works and (2) HA-NAS methods. Next, we present ablation studies to evaluate DNA's co-search algorithm and DAS engine. Finally, we visualize DNA's searched network and accelerator and discuss the insights.




\vspace{-0.2cm}
\subsection{Experiment setup} 
\label{sec:exp_setup}
\vspace{-0.1cm}
\subsubsection{Evaluation baselines and datasets} 
For evaluating DNA \underline{over SOTA co-exploration works}, we consider (1) three co-exploration FPGA baselines: HS-Co-Opt~\cite{jiang2019hardware}, BSW~\cite{abdelfattah2020best}, EDD~\cite{li2020edd} and (2) three co-exploration ASIC baselines: NASAIC~\cite{yang2020co}, NHAS~\cite{Lin2019NeuralHardwareAS}, and DANCE~\cite{choi2020dance}. For benchmarking \underline{over SOTA HA-NAS methods}, we consider four baselines: EfficientNet-B0~\cite{tan2019efficientnet}, FBNet~\cite{wu2019fbnet}, FBNet-V2~\cite{wan2020fbnetv2}, and ProxylessNAS~\cite{cai2018proxylessnas}. For \underline{evaluating DNA's DAS engine}, we consider both expert-designed and tool-generated SOTA accelerators in \cite{fpga_going_deepers, exploring_hetero, zhang2018dnnbuilder}. Our experiments consider three datasets: CIFAR-10, CIFAR-100, and ImageNet.


\begin{table*}[!t]
  \centering

  \caption{
Benchmark DNA over SOTA co-exploration works for generating both FPGA- and ASIC-based accelerators. Note that we quantize the DNA generated network to 8-bit when comparing with EDD~\cite{li2020edd} with a searched precision for a fair comparison.}
  \vspace{-0.3em}
    \resizebox{\linewidth}{!}{
    \begin{tabular}{ccccccccc}
    \toprule
    

    \textbf{Method target on}  & \multirow{2}{*}{\textbf{Dataset}} & \textbf{Network} & \textbf{Accelerator} & \textbf{Joint} & \textbf{Search Time} & \textbf{DSP} & \textbf{Accuracy} & \multirow{2}{*}{\textbf{FPS}} \\
    \textbf{FPGA} &  & \textbf{Space} &  \textbf{Space} & \textbf{Space} & \textbf{(GPU Hours)} & \textbf{Limit} & \textbf{ (\%)}& \\
    
    \midrule
    \midrule
    \multicolumn{1}{c}{HS-Co-Opt~\cite{jiang2019hardware}} & \multirow{2}{*}{CIFAR-10} & 1.15E+18 &   1     & \multicolumn{1}{c}{1.15E+18} & \multicolumn{1}{c}{103.9} & \multirow{2}{*}{450} & \multicolumn{1}{c}{85.19} & \multicolumn{1}{c}{35.5} \\
    
    \multicolumn{1}{c}{DNA (Proposed)} &  & 9.85E+20 & 2.24E+27 & \multicolumn{1}{c}{2.21E+48} & \multicolumn{1}{c}{\textbf{4.2 \textcolor{blue}{($\downarrow$24.7$\times$)}}} &   & \multicolumn{1}{c}{\textbf{96.10 \textcolor{blue}{($\uparrow$10.91)}}} & \multicolumn{1}{c}{\textbf{52.4 \textcolor{blue}{($\uparrow$1.48$\times$)}}} \\
    
    
    \midrule
    \multicolumn{1}{c}{BSW~\cite{abdelfattah2020best}} & \multirow{2}{*}{CIFAR-100} & 4.20E+05 & 8.64E+03 & \multicolumn{1}{c}{3.63E+09} & \multicolumn{1}{c}{5184} & \multirow{2}{*}{512} & \multicolumn{1}{c}{72.00} & \multicolumn{1}{c}{54.5} \\

    \multicolumn{1}{c}{DNA (Proposed)}  &  & 9.85E+20 & 2.24E+27 & \multicolumn{1}{c}{2.21E+48} & \multicolumn{1}{c}{\textbf{4.2 \textcolor{blue}{($\downarrow$1234.3$\times$)}}} &  & \multicolumn{1}{c}{\textbf{79.35 \textcolor{blue}{($\uparrow$7.35)}}} & \multicolumn{1}{c}{\textbf{64.3 \textcolor{blue}{($\uparrow$1.18$\times$)}}} \\
    
    \midrule

    \multicolumn{1}{c}{HS-Co-Opt~\cite{jiang2019hardware}} & \multirow{2}{*}{ImageNet} & 2.22E+18 &   1     & \multicolumn{1}{c}{2.22E+18} & \multicolumn{1}{c}{266.8} & \multirow{2}{*}{450} & \multicolumn{1}{c}{70.24} & \multicolumn{1}{c}{10.5} \\
    
    \multicolumn{1}{c}{DNA (Proposed)}  & & 9.85E+20 & 2.24E+27 & \multicolumn{1}{c}{2.21E+48} & \multicolumn{1}{c}{\textbf{144 \textcolor{blue}{($\downarrow$1.9$\times$)}}} &   & \multicolumn{1}{c}{\textbf{75.70 \textcolor{blue}{($\uparrow$5.46)}}} & \multicolumn{1}{c}{\textbf{31.9 \textcolor{blue}{($\uparrow$3.04$\times$)}}} \\
  
    \midrule

    \multicolumn{1}{c}{EDD~\cite{li2020edd}}  & \multirow{2}{*}{ImageNet} & 3.65E+19 &   -     & \multicolumn{1}{c}{-} & \multicolumn{1}{c}{-} & \multirow{2}{*}{900} &
    \multicolumn{1}{c}{74.40} & \multicolumn{1}{c}{40.2} \\
    
    \multicolumn{1}{c}{DNA 8-bit (Proposed)}  & & 9.85E+20 & 2.24E+27 & \multicolumn{1}{c}{2.21E+48} & \multicolumn{1}{c}{\textbf{144}} &
     & \multicolumn{1}{c}{\textbf{75.10 \textcolor{blue}{($\uparrow$0.7)}}} & \multicolumn{1}{c}{\textbf{78.1 \textcolor{blue}{($\uparrow$1.94$\times$)}}} \\
  
    \midrule
    \midrule

    
    \textbf{Method target on}  & \multirow{2}{*}{\textbf{Dataset}} & \textbf{Network} & \textbf{Accelerator} & \textbf{Joint} & \textbf{Search Time} & \textbf{Area} & \textbf{Accuracy} & \textbf{EDP} \\
     \textbf{ASIC} & & \textbf{Space} &  \textbf{Space} & \textbf{Space} & \textbf{(GPU Hours)} & \boldmath{ ($mm^2$)} & \textbf{ (\%)}& \textbf{(J * clock cycle)} \\

    \midrule
    \multicolumn{1}{c}{NASAIC~\cite{yang2020co}}  & \multirow{2}{*}{CIFAR-10} & 1.70E+03 & 9.84E+05 & \multicolumn{1}{c}{1.67E+09} & \multicolumn{1}{c}{4.6} & \multicolumn{1}{c}{3.34E+03} & \multicolumn{1}{c}{92.62} & \multicolumn{1}{c}{1.62E+06} \\

    \multicolumn{1}{c}{DNA (Proposed)}  &  & 9.85E+20 & 2.24E+27 & \multicolumn{1}{c}{2.21E+48} & \multicolumn{1}{c}{\textbf{4.2 \textcolor{blue}{($\downarrow$1.1$\times$)}}} &
    \multicolumn{1}{c}{\textbf{5.92E-01}} & \multicolumn{1}{c}{\textbf{96.50 \textcolor{blue}{($\uparrow$3.88)}}} & \multicolumn{1}{c}{\textbf{4.99E+03 \textcolor{blue}{($\downarrow$324.0$\times$)}}} \\
    
    \bottomrule
    \end{tabular}%
    }
  \label{tab:exp_space}
  \vspace{-1em}
\end{table*}
\subsubsection{Software experiment setup}

\textbf{Search and evaluation on CIFAR-10/100.} \underline{Search space:} we adopt the same search space as~\cite{wu2019fbnet}, except the stride settings for each group (i.e., a set of blocks with the same number of output channels), in order to adapt to the resolution of the input images in CIFAR-10/100. In particular, we follow the stride settings of MobileNetV2 on CIFAR-10/100 in~\cite{wang2019e2}, which is $[1, 1, 2, 2, 1, 2, 1]$ for all the seven groups.
\underline{Search settings:} the search for the optimal DNNs and accelerators adopts 50 epochs with a batch size of 64.  In particular, 
we (1) update the supernet weights on half of the training dataset using an SGD optimizer with a momentum of 0.9 and an initial learning rate of 0.025 associated with a cosine decay, and (2) update the network parameters on the other half of the training dataset using an Adam optimizer with a momentum of 0.9 and a fixed learning rate of 3E-4. Like SOTA differentiable NAS methods, 
we apply Gumbel-Softmax on the network parameters and treat each parameter option's weighted coefficients as their contribution to the supernet, following~\cite{wu2019fbnet}, where the initial temperature is set to 3 and decayed by 0.92 at each epoch. Meanwhile, for updating the hardware accelerator parameters, we conduct updates at each epoch using an SGD optimizer with a momentum of 0.9 and a fixed learning rate of 1E-9. 
\underline{Evaluate the derived networks:} for training the derived networks from scratch, we follow~\cite{liu2018darts} and adopt an SGD optimizer with a momentum of 0.9 and an initial learning rate of 0.01 associated with a cosine decay. Each network is trained for 600 epochs with a batch size of 96.

\textbf{Search and evaluation on ImageNet.} \underline{Search space:} we adopt the same search space as~\cite{wu2019fbnet} which is a SOTA search space for generating efficient DNNs via NAS.
\underline{Search settings:} we follow the same hyper-parameter settings as ~\cite{wu2019fbnet} for searching on ImageNet, while additionally updating the hardware accelerator parameters at each epoch using an SGD optimizer with a momentum of 0.9 and a fixed learning rate of 1E-9. Specifically, the search for both the optimal network and accelerator adopts 90 epochs with each having a batch size of 192; we first update the supernet's weights on 80\% of the training dataset using an SGD optimizer with a momentum of 0.9 and an initial learning rate of 0.1 associated with a cosine decay, and then update the network structure parameters on the remaining 20\% of the training dataset using an Adam optimizer with a momentum of 0.9 and a fixed learning rate of 1E-2. The initial temperature of Gumbel-Softmax is set to 5 and then decayed by 0.956 at each epoch.
\underline{Evaluate the derived networks:} we train the derived networks using an SGD optimizer with a momentum of 0.9 and an initial learning rate of 0.05 associated with a cosine decay. Each network is trained for 180 epochs with a batch size of 512.

\textbf{Hardware cost and hyperparameters.} We adopt Frame-Per-Second (FPS), latency, or Energy-Delay-Product (EDP) as the hardware-cost loss (i.e., $L_{hw}$ in Eq.~\ref{eq:update_alpha}) to compare with various baselines and set $M=10$ for Eq.~\ref{eqn:expect} in all the experiments. We empirically find that $M$ can trade-off the search time and stability, i.e., smaller $M$ reduces the search time while its resulting accelerators might be sub-optimal (e.g., their efficiency does not meet the specification). Our observation is that the performance of our DNA is not sensitive to $M$ for $M>5$, and we thus choose $M=10$.

\subsubsection{Hardware experiment setup} 
\label{hardware-setup}
\textbf{Accelerator evaluation methodology.} 
To evaluate DNA's generated accelerators, we adopt standard FPGA and ASIC evaluation and implementation flows. 
For evaluating FPGA-based accelerators, we employ the Vivado HLS design flow~\cite{vivado_HLS}. Note that during the search for FPGA-based accelerators, DNA makes use of a SOTA accelerator performance predictor~\cite{xu2020autodnnchip} to obtain fast and reliable estimation. For evaluating ASIC-based accelerators, we use SOTA accelerator performance estimation tools Timeloop~\cite{parashar2019timeloop} and Accelergy~\cite{wu2019accelergy} during and after the search. As \cite{parashar2019timeloop,wu2019accelergy} suggest, the unit energy and latency are obtained using CACTI7~\cite{cacti7} and Aladdin~\cite{aladdin} based on a commercial 32nm or 45nm CMOS technology depending on that of the baselines for a fair comparison.

\begin{figure*}[htp]
    \centering
    \includegraphics[width=\textwidth]{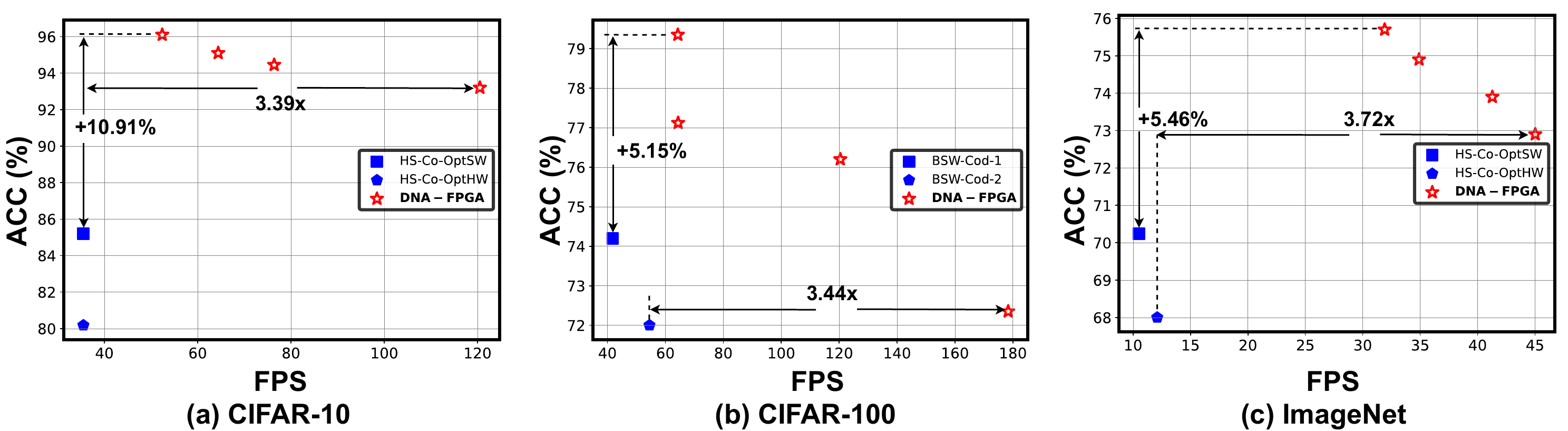}
        \vspace{-1.6em}
    \caption{DNA generated FPGA-based accelerators over those of SOTA co-exploration works HS-CO-Opt~\cite{jiang2019hardware} and BSW~\cite{abdelfattah2020best}, where we adopt the same DSP limits as the baselines, i.e., 450/512/450 on CIFAR-10/100/ImageNet, respectively. }
    \label{fig:fpga_dnacos}
    \vspace{-1.5em}
\end{figure*}

 \vspace{-0.5em}
\subsection{DNA over SOTA co-exploration works}
 \label{sec:DNA_evaluation}
\vspace{-0.2em}

\textbf{Search efficiency.} Here we benchmark our DNA over SOTA co-exploration works~\cite{jiang2019hardware, abdelfattah2020best, li2020edd, yang2020co} in terms of the search space size and search time. As shown in Tab.~\ref{tab:exp_space}, DNA can handle a remarkably larger (e.g., 2.21E+48 vs. 3.63E+09) joint search space while requiring the shortest search time (e.g., 4.2 vs. 5184 GPU hours), as compared to all the SOTA baselines. Specifically, \underline{on ImageNet} DNA achieves a 5.46\% and 0.7\% higher accuracy with a 3.04$\times$ and 1.94$\times$ higher FPS under a 450 and 900 Digital-Signal-Processor (DSP) limit, respectively, as compared to HS-Co-Opt~\cite{jiang2019hardware} and EDD~\cite{li2020edd}; \underline{on CIFAR-100} DNA achieves a 1234.3$\times$ speed-up in search time when handling a 6.1E+38$\times$ larger search space, resulting in a 7.35\% better accuracy and 1.18$\times$ higher FPS. 
This set of evaluations validates that DNA's differentiable co-search can indeed handle a notably larger joint space with much improved search efficiency, enabling both fast development and superior performance.

\begin{figure}[!b]
    \centering
    \vspace{-2em}
    \includegraphics[width=0.8\linewidth]{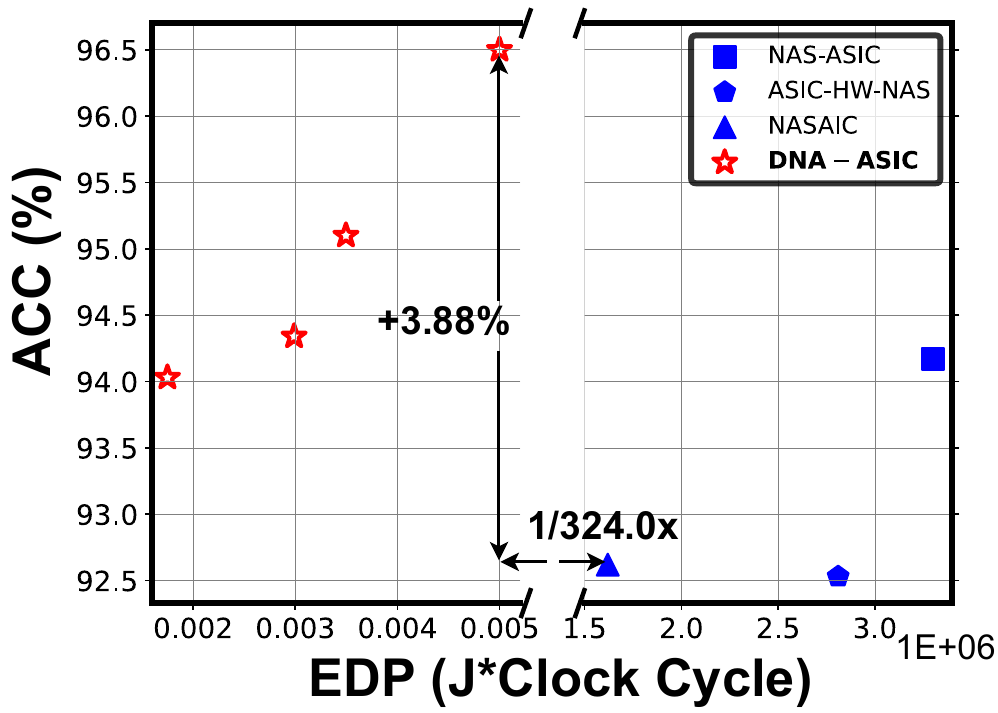}
  \vspace{-1em}
    \caption{
     Accuracy vs. EDP of DNA generated ASIC-based accelerators over three SOTA co-exploration designs on CIFAR-10 in NASAIC~\cite{yang2020co}.
    }
    \label{fig:ASIC_dnacos}
    \vspace{-1em}
\end{figure}

\textbf{Achieved FPS on FPGA.}
Here we compare DNA's generated accelerators with those of the co-exploration baselines, HS-Co-Opt~\cite{jiang2019hardware} and BSW~\cite{abdelfattah2020best}, in terms of the achieved FPS and accuracy under the same FPGA resource budgets and datasets, as shown in Fig.~\ref{fig:fpga_dnacos}. 
We can make two observations from this evaluation. First, DNA generated accelerators consistently achieve the highest FPS with the same or even a higher accuracy, validating DNA's advantage.
Specifically, \underline{on CIFAR-10} with a $450$ DSP limit, DNA boosts the FPS by 1.48$\times$ $\sim$ 3.39$\times$, while offering a 8.01\% $\sim$ 10.91\% higher accuracy over HS-Co-Opt; \underline{on CIFAR-100} with a $512$ DSP limit, DNA achieves a 1.18$\times$ $\sim$ 3.44$\times$ higher FPS while boosting the accuracy by 0.35\% $\sim$ 5.15\% over BSW; and 
\underline{on ImageNet} with a $450$ DSP limit, DNA achieves a $3.04\times$ higher FPS with a  $5.46\%$ higher accuracy as compared to HS-Co-Opt. 
Second, under various resource settings and datasets, DNA's generated accelerators can trade-off the achieved FPS and accuracy (e.g., a higher accuracy at the cost of a reduced FPS) under the given FPGA resource budget. 

\textbf{Achieved EDP/latency on ASIC.}
For evaluating DNA generated ASIC-based accelerators, we consider three SOTA co-search works, NASAIC~\cite{yang2020co}, NHAS ~\cite{Lin2019NeuralHardwareAS}, and DANCE~\cite{choi2020dance}, based on their reported metrics and the same settings (e.g., precision, area, and dataset). Fig.~\ref{fig:ASIC_dnacos} shows the benchmarking \underline{over NASAIC}, from which we can make two observations. First, DNA generated accelerators achieve a much improved EDP while leading to a higher accuracy as compared with NASAIC, e.g., 324.0$\times$ EDP reduction together with a 3.88\% improvement in accuracy. Second, similar to the experiments for designing FPGA-based accelerators, we can see that DNA generated ASIC-based accelerators can flexibly trade-off between the achieved accuracy and efficiency (i.e., EDP here). Note that the surprisingly higher EDP and area for NASAIC, as noted in Tab.~\ref{tab:exp_space}, is caused by the much reduced hardware utilization due to their design consideration of heterogeneous tasks, leading to severely low utilization when executing one task. Tab.~\ref{tab:asic} summarizes the evaluation results \underline{over NHAS}, where DNA adopts 
the same dataset (ImageNet), network precision (4-bit), and accelerator metric (latency) as the NHAS baseline. We can see that DNA generated accelerators outperform NHAS in all aspects: a $0.96\%$ higher accuracy on ImageNet while having a $20.89\%$ and $6.3\%$ reduction in the required latency and area, respectively. Additionally, we benchmark DNA over the reported results of DANCE~\cite{choi2020dance} under the same settings, where a 16-bit network precision is adopted to be comparable with DANCE. As shown in Tab.~\ref{tab:asic}, DNA achieves both better hardware performance and higher accuracy, with a $3.50\%$ higher accuracy on ImageNet while having a $64.94\%$ and $22.34\%$ reduction in the latency and area, respectively.

\begin{table}[!b]
  \vspace{-1.3em}
  \centering
  \caption{Evaluating DNA generated ASIC-based accelerators over NHAS~\cite{Lin2019NeuralHardwareAS} and DANCE~\cite{choi2020dance} under the same setting.
}
\vspace{-0.8em}
    \resizebox{0.49\textwidth}{!}{
\begin{tabular}{ccccc}
\hline
\textbf{Optimization} & \textbf{Accuracy} & \textbf{Latency} & \textbf{Area} & Precision \\
\textbf{Methods} & \textbf{(\boldmath{$\%$)}} & \textbf{\boldmath{($ms$)}} & \textbf{\boldmath{($mm^2$)}} &\textbf{\boldmath{($bit$)}}  \\\hline \hline
NHAS~\cite{Lin2019NeuralHardwareAS} & 70.74 & 1.58 & 5.87 & 4 \\
DNA (Proposed) & \textbf{71.70 \textcolor{blue}{($\uparrow$0.96)}} & \textbf{1.25 \textcolor{blue}{($\downarrow$20.89\%)}} & \textbf{5.50 \textcolor{blue}{($\downarrow$6.3\%)}} & 4\\
DANCE~\cite{choi2020dance} & 68.70 & 8.13 & 2.73 & 16 \\
DNA (Proposed) & \textbf{72.20 \textcolor{blue}{($\uparrow$3.50)}} & \textbf{2.85 \textcolor{blue}{($\downarrow$64.94\%)}} & \textbf{2.12 \textcolor{blue}{($\downarrow$22.34\%)}} & 16\\
\hline
\end{tabular}
    }
  \label{tab:asic}%
  \vspace{-1em}
\end{table}%

\vspace{-0.2em}
\subsection{DNA ablation: DNA over SOTA HA-NAS works}
\begin{figure}[!ht]
    \centering
    \includegraphics[width=1\linewidth]{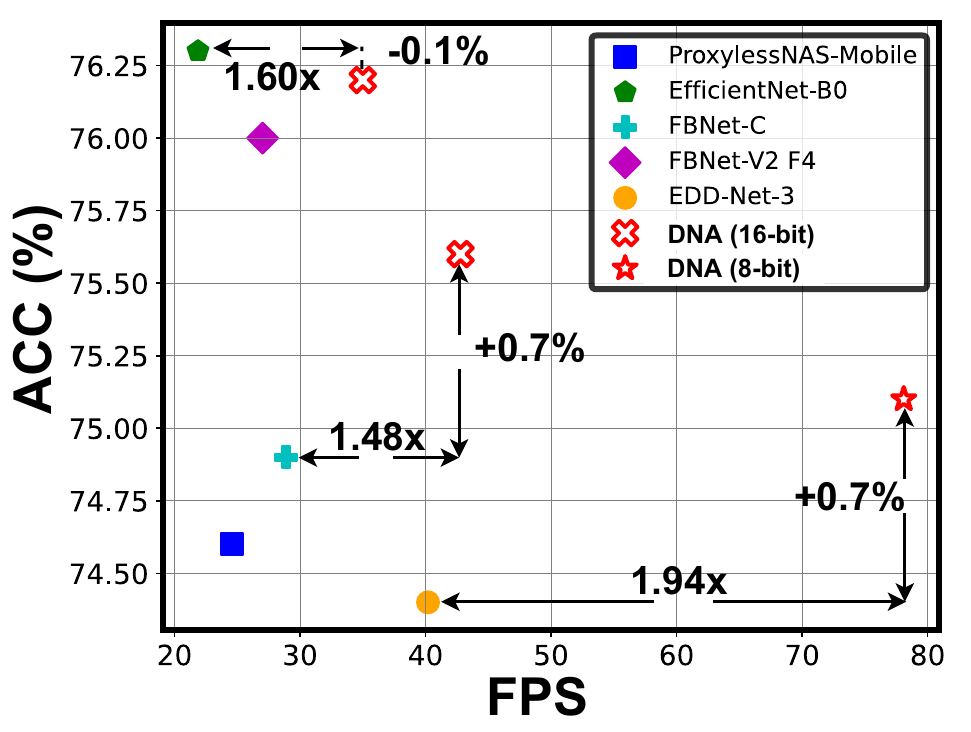}
    \vspace{-1.2em}
    \caption{Accuracy vs. FPS of DNA and SOTA HA-NAS generated networks~\cite{tan2019efficientnet, wu2019fbnet, wan2020fbnetv2, cai2018proxylessnas} and EDD~\cite{li2020edd}.}
    \label{fig:benchmark_nas}
    \vspace{-1em}
\end{figure}

\vspace{-0.2em}
\label{sec:exp_cosearch}
In this sub-section, we evaluate DNA over SOTA HA-NAS works. Specifically, we compare the acceleration efficiency of DNA generated accelerators and SOTA HA-NAS generated networks when they are accelerated by their optimal accelerators, which are generated by the DAS engine under a DSP constraint of 900 (the maximum one in a ZC706 board~\cite{zc706}) to maximize the achieved FPS, as summarized in Fig.~\ref{fig:benchmark_nas}. Note that we consider two kinds of precision, i.e., 16-bit and 8-bit, and we adopt 8-bit for a fair comparison over EDD with searched precision and adopt 16-bit when comparing with all other HA-NAS works to maintain their reported accuracy. We can see that (1) compared with EDD in searched precision, DNA (8-bit) achieves a 0.7\% higher accuracy and 1.94x higher FPS; and (2) compared with SOTA HA-NAS methods, DNA (16-bit) achieves a 1.48$\times$ higher FPS while having a +0.7\% higher accuracy over FBNet and a 1.60$\times$ higher FPS with comparable accuracy (-0.1\%) over EfficientNet-B0.

\subsection{DNA ablation: co-search algorithm}
\label{sec:exp_cosearch}

\textbf{Experiment setting.} To evaluate our DNA's effectiveness and necessity, we consider sequential search (SEQ-Opt) on the same search space and MobileNetV2~\cite{sandler2018mobilenetv2} optimized by DAS as our baseline. For the former, we first use DNA's DNS engine to search for the optimal DNN structure that minimizes the task loss and a theoretical computational loss, i.e., FLOPs, and then adopt DNA's DAS engine to search for the corresponding optimal accelerator of the resulting DNN from the previous step. For the latter, the FPS/EDP of MobileNetV2 are obtained from the optimal accelerator searched by DNA's DAS engine. For a fair comparison, we ensure that the DSP/ASIC area is the same for all experiments.


\begin{figure}[!t]
    \centering
    \includegraphics[width=\linewidth]{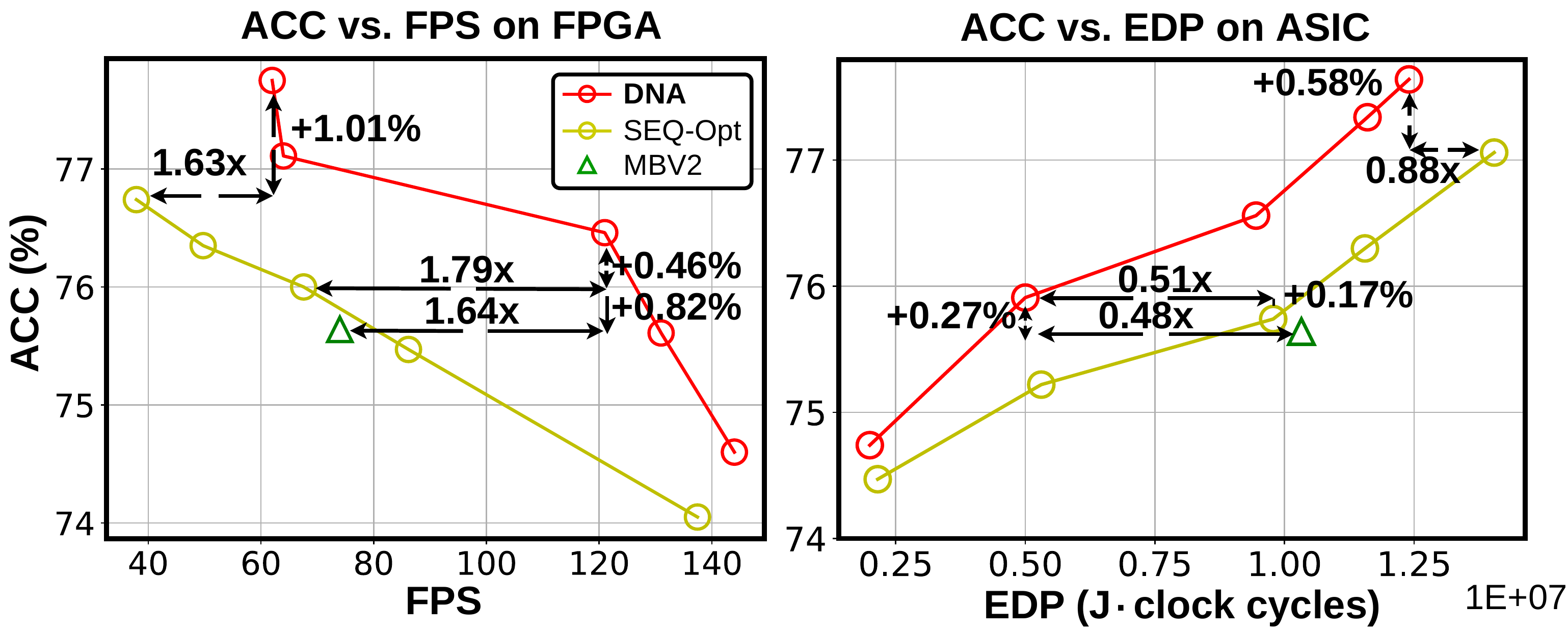}
    \vspace{-1.5em}
    \caption{Accuracy vs. FPS/EDP of DNA's co-search algorithm, sequential search (SEQ-Opt) and MobileNetV2 (MBV2) optimized by DAS evaluated on CIFAR-100.}
    \label{fig:acc_edp_trade_off}
      \vspace{-1em}
\end{figure}

\textbf{Results.} Fig.~\ref{fig:acc_edp_trade_off} shows the achieved accuracy vs. FPS/EDP on both FPGA/ASIC, indicating that DNA's co-search algorithm consistently outperforms the baselines. In particular, for the case on FPGA, we can see that (1) DNA outperforms MobileNetV2 and its optimal accelerator with a 1.64$\times$ improvement in FPS and a 0.82\% higher accuracy; and (2) DNA achieves a 1.63$\times$ improvement and a 1.01\% higher accuracy over the SEQ-Opt baseline. For the case on ASIC, we can make the observations that (1) DNA outperforms MobileNetV2 and its optimal accelerator with a 0.51$\times$ reduced EDP and a 0.27\% higher accuracy; and (2) DNA achieves a 0.88$\times$ reduced EDP and a 0.58\% higher accuracy over the SEQ-Opt baseline. This set of experiments validate both the necessity of co-search and the effectiveness of DNA's co-search algorithm.


\begin{table}[!b]
\centering
\vspace{-1em}
\caption{Evaluating DNA over random network and accelerator search on CIFAR-100 evaluated on FPGA.}
\vspace{-0.5em}
\resizebox{0.5\textwidth}{!}
{
\begin{tabular}{cccc}
\hline
\textbf{Search Strategy} & \textbf{Accuracy (\%)} & \textbf{FPS} \\

\hline
\hline
random network/accelerator search & 70.3 - 75.1 & 31.9 - 82.5 \\
DNA (Proposed) & \textbf{74.6 - 78.1} &  \textbf{64.7 - 158.6} \\
\hline
\end{tabular}
}
\label{tab:random_search}

\end{table}

\begin{table*}[bth]
  \centering
  \vspace{-0.2em}
    \caption{Our DAS generated FPGA accelerators vs. SOTA FPGA accelerators on a Zynq XC70Z45 FPGA, when accelerating SOTA networks (VGG16 and AlexNet) on ImageNet at a frequency of 200MHz.}
  \vspace{-0.3em}
    \resizebox{\linewidth}{!}{
        \begin{tabular}{ccccccc}
        \hline
 & \cite{zhang2018dnnbuilder} & \cite{exploring_hetero}&\cite{fpga_going_deepers} &\textbf{DAS generated } & \cite{zhang2018dnnbuilder} & \textbf{DAS generated}\\ \hline
 \hline
         \textbf{Network}   & VGG16 & VGG16& VGG16 & VGG16&  AlexNet& AlexNet \\
        \textbf{Resource Utilization} & 680/900 DSP & 824/900 DSP& 780/900 DSP& 723/900 DSP& 808/900 DSP& 704/900 DSP \\ \hline \hline
        \textbf{Performance (GOP/s)}  & 262 & 230& 137 & \textbf{291 \textcolor{blue}{($\uparrow$1.11$\times$ - $\uparrow$2.12$\times$)}}& 247 & \textbf{272 \textcolor{blue}{($\uparrow$1.10$\times$)}} \\ \hline
        \end{tabular}        }
  \label{tab: dhs_ablation_fpga}
  \vspace{-1em}
\end{table*}

We also evaluate DNA's co-search algorithm over a random search by comparing their searched accelerators in the same joint search space, as summarized in Tab.~\ref{tab:random_search}. Specifically, we first conduct random network search by randomly sampling 300 DNNs, then perform random accelerator search by randomly sampling 300 accelerators for each network, and finally select the DNNs and paired accelerators that together achieve the best accuracy-FPS trade-offs. The observation is that random network search tends to result in small networks (higher density in the adopted SOTA network space) with inferior accuracy, while the random accelerator search cannot find an efficient accelerator even for small networks. 

\begin{figure*}[!t]
    \centering
    \vspace{1.em}
    \includegraphics[width=\linewidth]{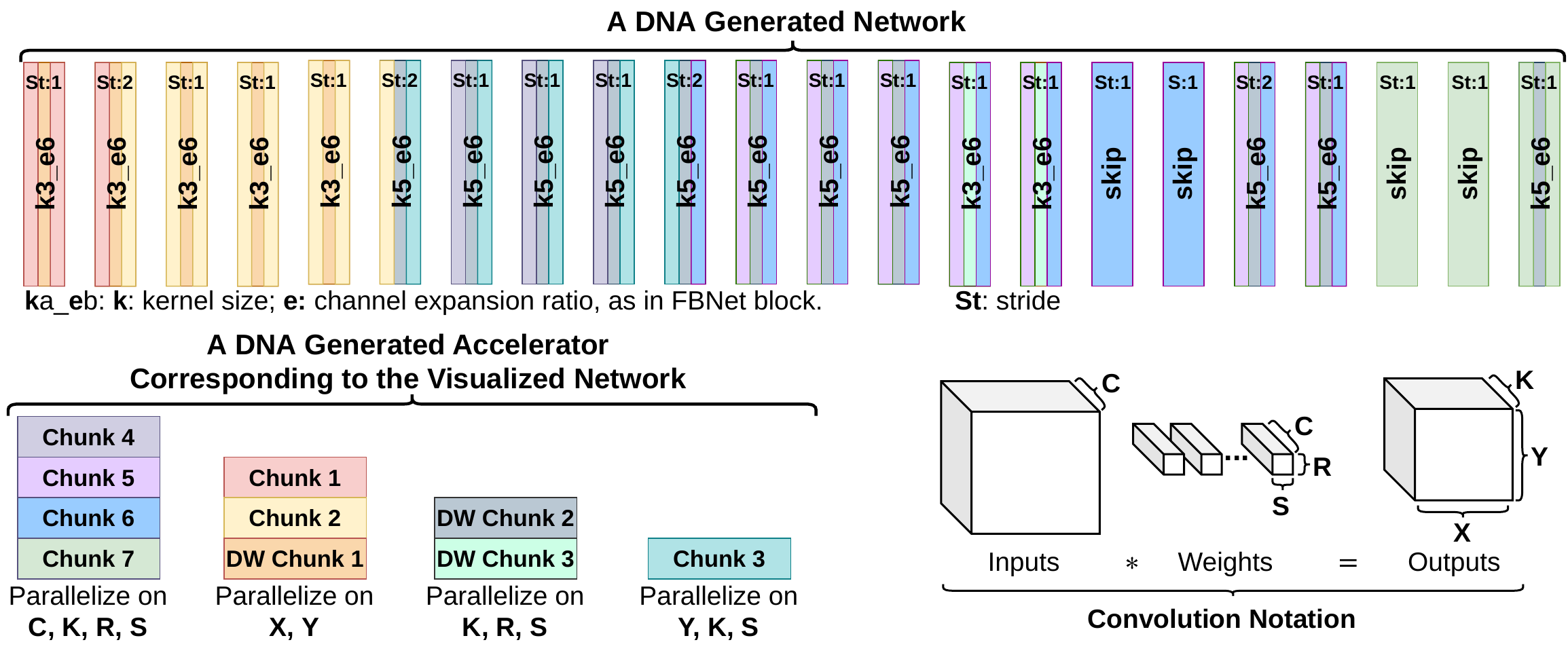}
    \vspace{-2.0em}
    \caption{Visualizing a DNA generated network and accelerator, which achieves a 75.6\% accuracy on ImageNet and 42 FPS on a ZC706 FPGA, where the block definition follows~\cite{wu2019fbnet} with \textbf{k3\_e6} denoting the building block associated with a kernel size of 3 and a channel expansion ratio of 6. X/Y, R/S, and C/K denote the dimensions of a convolution (see the bottom right figure), i.e., output feature width/height, kernel height/width, and input/output channel, respectively.
    }
    \label{fig:vis}
     \vspace{-1.5em}
\end{figure*}

\vspace{-0.2em}
\subsection{DNA ablation: DAS over SOTA DNN accelerators}
\label{sec:exp_dhs}
\vspace{-0.2em}

The proposed DAS is one key enabler of our DNA framework. To evaluate its efficacy, we compare the hardware efficiency of the DAS generated accelerators with SOTA accelerators under the same datasets and models. 
We consider three representative FPGA-based accelerators including ~\cite{fpga_going_deepers,exploring_hetero,zhang2018dnnbuilder}, when accelerating two networks (AlexNet and VGG16) on ImageNet. 
The results in Tab.~\ref{tab: dhs_ablation_fpga} show that our DNA generated accelerators outperform \textbf{both expert-designed and tool-generated} SOTA accelerators under the same dataset, DNNs, and FPGA resources. For example, 
DAS generated accelerators achieve up to $2.12 \times$ improvement in throughput on VGG16 under the same setting. The consistent better performance of our DAS generated accelerators validates the effectiveness of our DAS in navigating the large and discrete design space of DNN accelerators to search for optimal DNN accelerators. Note that when using DAS to generate optimal accelerators, we adopt the same precision and FPGA resources as the baselines for a fair comparison.

\subsection{Visualizing DNA's searched network and accelerator}
To better understand the superior performance of DNA generated networks and accelerators, Fig.~\ref{fig:vis} visualizes a DNA-generated network and accelerator on ImageNet, corresponding to the 16-bit DNA in Fig.~\ref{fig:benchmark_nas} where it achieves an accuracy of 75.6\% and a FPS of 42 under an FPGA DSP limit of 900 (the maximum one in a ZC706 board~\cite{zc706}). First, we can see that the resulting network tends to select skip connections and adopt larger channel numbers, i.e., it trades the depth for the width to ensure both higher accuracy and hardware efficiency when being executed on the searched accelerator. Second, on the accelerator side, to maximize the potential throughputs, we observe that the generated accelerator employs a chuck-based pipeline micro-architecture, i.e., the whole network is partitioned and assigned to multiple pipelined chunks (sub-accelerators) in a non-consecutive manner which can be better tailored to optimize different layers' specific dimensions as elaborated in~\cite{shen2017ISCA}.
Furthermore, we observe that the chunks for the early layers in the network tend to parallelize more in the feature height/width dimensions, while the deeper layers parallelize the input/output channel dimensions, adapting to the dimensions which have 
more parallelism opportunities. 

\section{Conclusion}
   
We propose DNA, a generic differentiable network-accelerator co-search framework for automatically searching for matched DNN structures and accelerators to maximize both task accuracy and hardware efficiency.  
DNA is made possible by: (1) a new co-search algorithm that can efficiently navigate over the prohibitively large joint network and accelerator search space to enable simultaneous search of DNNs' structures and accelerators; and
 (2) a generic differentiable accelerator search engine which integrates gradient-based optimization and our constructed generic accelerator search space to handle the large and discrete accelerator space. 
Extensive experiments and ablation studies based on FPGA measurements and ASIC synthesis validate that DNA generated networks and accelerators consistently outperform all the SOTA baselines in terms of both task accuracy and hardware efficiency (even better than expert-designed ones yet requiring a much reduced development time), while notably boosting the search efficiency. As such, DNA promises to greatly close the gap between rapidly growing DNNs and DNN accelerators' slow development.
   

\nocite{langley00}
\bibliography{ref}
\bibliographystyle{mlsys2020}



\end{document}


\twocolumn[
\mlsystitle{\PaperTitle}



\mlsyssetsymbol{equal}{*}

\begin{mlsysauthorlist}
\mlsysauthor{Aeiau Zzzz}{equal,to}
\mlsysauthor{Bauiu C.~Yyyy}{equal,to,goo}
\mlsysauthor{Cieua Vvvvv}{goo}
\mlsysauthor{Iaesut Saoeu}{ed}
\mlsysauthor{Fiuea Rrrr}{to}
\mlsysauthor{Tateu H.~Yasehe}{ed,to,goo}
\mlsysauthor{Aaoeu Iasoh}{goo}
\mlsysauthor{Buiui Eueu}{ed}
\mlsysauthor{Aeuia Zzzz}{ed}
\mlsysauthor{Bieea C.~Yyyy}{to,goo}
\mlsysauthor{Teoau Xxxx}{ed}
\mlsysauthor{Eee Pppp}{ed}
\end{mlsysauthorlist}

\mlsysaffiliation{to}{Department of Computation, University of Torontoland, Torontoland, Canada}
\mlsysaffiliation{goo}{Googol ShallowMind, New London, Michigan, USA}
\mlsysaffiliation{ed}{School of Computation, University of Edenborrow, Edenborrow, United Kingdom}

\mlsyscorrespondingauthor{Cieua Vvvvv}{c.vvvvv@googol.com}
\mlsyscorrespondingauthor{Eee Pppp}{ep@eden.co.uk}

\mlsyskeywords{Machine Learning, MLSys}

\vskip 0.3in

]



\printAffiliationsAndNotice{\mlsysEqualContribution} 

\begin{figure}[!t]
\begin{minipage}{0.48\textwidth}
    \begin{algorithm}[H]
        \label{alg:2dconv}
        \caption{CONV described using standard \textit{for-loop} description}
        \label{alg:2dconv}
        \begin{algorithmic}[1]
        \vspace{1pt} 
        \STATE{ \textit{for} $c=0: C-1$  \textbf{channel\_in}}
        \vspace{9pt}
        \STATE{\hspace{8pt}  \textit{for} $k=0: K-1$ \textbf{channel\_out}}
        \vspace{10pt}
        \STATE{\hspace{16pt}  \textit{for} $y=0: Y-1$ \textbf{output\_row}}
        \vspace{9pt}
        \STATE{\hspace{24pt}  \textit{for} $x=0: X-1$ \textbf{output\_col}}
        \vspace{9pt}
        \STATE{\hspace{32pt}  \textit{for} $r=0: R-1$ \textbf{kernel\_row}}
        \vspace{9pt}
        \STATE{\hspace{40pt}  \textit{for} $s=0: S-1$ \textbf{kernel\_col}}
        \vspace{12pt}
         \STATE{\hspace{48pt}\color{orange} // MAC operation}
        \STATE{\hspace{48pt}  ofmap[k][y][x]+=
         \\\hspace{48pt} ifmap[c][y+r][x+s]*kernel[c][k][r][s]}

        \end{algorithmic}
    \end{algorithm}
\end{minipage}
\end{figure}

\begin{figure}[!t]
\begin{minipage}{0.48\textwidth}
    \begin{algorithm}[H]
        \label{alg:2dconv2}
        \caption{CONV via nested \textit{for-loop} description with both \textit{\textit{parallel-for}}s and \textit{inner-loops} }
        \label{alg:2dconv2}
        \begin{algorithmic}[1]
        \STATE{\hspace{0pt}  {{\color{orange}// DRAM level }} }
        \vspace{-2pt}
        \STATE{... }
        \vspace{-2pt}
        \STATE{\hspace{0pt}  {{\color{orange}// Global buffer level }} }
        \STATE{ \textit{for} $c_2=0: C_2-1$  \textbf{channel\_in}}
        \vspace{-2pt}
        \STATE{\hspace{8pt}...}
        \vspace{-2pt}
        \STATE{\hspace{16pt}  \textit{for} $s_2=0: S_2-1$ \textbf{kernel\_col}}
        
        \vspace{4pt}
        \STATE{\hspace{0pt}  {{\color{orange}// NoC level}} }
        \STATE{ \textit{\textit{parallel-for}} $r_1=0: R_1-1$  \textbf{kernel\_row}}
        \STATE{ \textit{\textit{parallel-for}} $s_1=0: S_1-1$  \textbf{kernel\_col}}
        \vspace{4pt}
        
        \STATE{\hspace{0pt}  {{\color{orange}// RF level}} }
        \STATE{ \textit{for} $c_0=0: C_0-1$  \textbf{channel\_in}}
        \STATE{\hspace{8pt}...}

        \STATE{\hspace{16pt}  \textit{for} $s_0=0: S_0-1$ \textbf{kernel\_col}}
        \STATE{\hspace{24pt}  MAC operation}

        \end{algorithmic}
    \end{algorithm}
\end{minipage}
\end{figure}

\section{The nested \textit{for-loop} representation adopted in GADS}
\label{sec:for-loop}
Generally, the execution of a convolution operation can be described as a standard nested \textit{for-loop} description~\cite{chen2016eyeriss,parashar2019timeloop,blocking_cnn,zhang2015optimizing,zhao2020icassp} as shown in Alg.~\ref{alg:2dconv}, which iterates all the six related dimensions (annotated in bold) and calculates the element-wise outputs. 
For describing DNNs' computation flow in real accelerators, common practice is to adopt the nested \textit{for-loop} description in Alg.~\ref{alg:2dconv2} with additional primitives, including (1) \textit{memory-hierarchy} and (2) \textit{parallel-for}~\cite{zhao2020icassp}, 
thanks to its compatibility with standard DNN \textit{for-loop} description in Alg.~\ref{alg:2dconv} and intuitive representation. As such, our proposed GADS follows the same convention for ease of generalization and adoption. We here elaborate the aforementioned two additional primitives below:

\textbf{\textit{memory-hierarchy}}: SOTA accelerators adopt various memory hierarchies for maximizing data reuses and thus acceleration efficiency, motivating this primitive. The description in Alg.~\ref{alg:2dconv2} adopts 
multiple levels of nested loops with each level denoting one hierarchy memory to represent different memory hierarchies. For instance, as shown in Alg.~\ref{alg:2dconv2}, the standard \textit{for-loop} description is extended to multiple levels of loops with each level representing (1) temporal data computation and (2) movements of intermediate results within each memory hierarchy, e.g., $C$ is extended to $C_0$ and $C_2$ for tiling the data between RF and global buffer.

\textbf{\textit{parallel-for}}: the accelerators' parallel computation along a certain data dimensions. For instance, in Alg.~\ref{alg:2dconv2}, computation on different rows and columns for the kernels are distributed into $R1*S1$ PEs and executed in parallel.

\nocite{langley00}

\bibliography{ref}
\bibliographystyle{mlsys2020}

